\pdfoutput=1

\documentclass[11pt]{article}

\usepackage[final]{acl}

\usepackage{times}
\usepackage{latexsym}

\usepackage[T1]{fontenc}

\usepackage[utf8]{inputenc}

\usepackage{microtype}

\usepackage{inconsolata}

\usepackage{graphicx}

\usepackage{float} 

\title{Understanding Verbatim Memorization in LLMs \\ Through Circuit Discovery}

\author{
  \textbf{Ilya Lasy},
  \textbf{Peter Knees},
  \textbf{Stefan Woltran}
\\
\\    
  Faculty of Informatics, TU Wien, Vienna, Austria
\\
   Correspondence: \href{mailto:ilya.lasy@tuwien.ac.at}{ilya.lasy@tuwien.ac.at}
}

\begin{document}
\maketitle
\begin{abstract}
Underlying mechanisms of memorization in LLMs---the verbatim reproduction of training data---remain poorly understood. What exact part of the network decides to retrieve a token that we would consider as start of memorization sequence? How exactly is the models' behaviour different when producing memorized sentence vs non-memorized? In this work we approach these questions from mechanistic interpretability standpoint by utilizing transformer circuits---the minimal computational subgraphs that perform specific functions within the model. Through carefully constructed contrastive datasets, we identify points where model generation diverges from memorized content and isolate the specific circuits responsible for two distinct aspects of memorization. We find that circuits that initiate memorization can also maintain it once started, while circuits that only maintain memorization cannot trigger its initiation. Intriguingly, memorization prevention mechanisms transfer robustly across different text domains, while memorization induction appears more context-dependent.\textsuperscript{1}
\end{abstract}

\section{Introduction}

\footnotetext[1]{Code and data available at: \url{https://github.com/ilyalasy/memorization_circuits}}

Large Language Models (LLMs) have demonstrated remarkable capabilities across a broad spectrum of applications \citep{openai2024gpt4, deepseekai2025, llama2023}. Despite these impressive advancements, researchers have identified various challenges with these models, with memorization emerging as an important issue. Memorization in LLMs refers to the model's tendency to store and reproduce exact phrases or passages from its training data when prompted with appropriate contexts.

Memorization capabilities have both positive and negative aspects. On one hand, deliberate memorization enables models to store facts, concepts, and general knowledge that enhance their performance in tasks like question answering and information retrieval \citep{ranaldi2023pre, chen2023beyond, lu2024scaling}. On the other hand, undesirable memorization creates several problems: privacy risks when models expose personal information, security issues when they reveal passwords or credentials, copyright concerns when they reproduce protected content, and biases reflecting their training data. 
Furthermore, memorization complicates model transparency and interpretability, making it difficult to determine whether specific outputs reflect generalization or memorized content.

Interpretability research has provided important insights into memorization in LLMs, revealing that memorization is distributed across model layers with distinct patterns: early layers promote tokens in the output distribution, while upper layers amplify confidence \citep{haviv2023understanding}. Studies have demonstrated that memorized information shows distinct gradient patterns in lower layers and is influenced by attention heads focusing on rare tokens \citep{stoehr2023localizing}. However, there exists a promising approach for understanding neural networks that has yet to be fully applied to memorization: the circuits framework \citep{olah2020zoom, elhage2021mathematical}. This framework, which seeks to reverse-engineer model behavior by localizing it to subgraphs of the model's computation graph, has gained significant traction in interpretability research by providing mechanistic explanations of how models accomplish specific tasks \citep{conmy2023towards, wang2022interpretability}. Most state-of-the-art automatic circuit discovery algorithms rely on patching techniques, which require contrastive datasets with clean and corrupted examples \citep{conmy2023towards, syed2023attribution, hanna2023edge}. Creating such datasets is challenging for open-ended tasks like memorization. As a result, circuit discovery has mostly been applied to small, well-defined tasks like indirect object identification (IOI) \citep{wang2022interpretability}, greater-than comparisons \citep{hanna2023does}, or factual knowledge recall \citep{knowledge_circuits}. These simpler tasks make circuit discovery more tractable, but leave open the question of how to apply similar techniques to understand more complex behaviors like memorization.

In this paper, we introduce an approach to studying memorization circuits in language models trained on The Pile dataset. 
We focus specifically on the Wikipedia subset of The Pile, as its clean, well-structured content makes it easier to analyze when manually creating our contrastive datasets. 
Using this data, we identify highly memorized sentences and precisely locate the divergence points where the model transitions from memorization to generation.
Our contributions include:
\begin{itemize}
    \item Creating two distinct datasets addressing different aspects of memorization: one for the initial decision to retrieve memorized content and another for continuing along memorized paths
    \item Discovering compact circuits ($ \leq5\%$ model edges) of these two tasks using attribution patching techniques.
    \item Evaluating faithfulness of detected subgraphs in different experimental settings to ensure our circuits reliably capture memorization mechanisms — including both blocking memorization (corrupt-to-clean patching) and inducing memorization (clean-to-corrupt patching), testing how circuits generalize between different memorization tasks, and validating circuit performance across different text domains (\S \ref{sec:experiments})
    \item Demonstrating that circuits that can trigger memorization can also maintain it once started, while circuits that only maintain memorization cannot trigger its start (\S \ref{sec:results:cross-task})
    \item Finding that circuits that prevent memorization work across different text domains of The Pile (GitHub code, Enron Emails, and Common Crawl), while circuits that cause memorization are more specific to each domain (\S \ref{sec:results:cross-corpus})  

\end{itemize}
Our findings help us better understand how memorization works in language models and may lead to improved ways to control this behavior.
\section{Related Work}

\subsection{Memorization in Language Models}

\citet{carlini2019extracting} introduced the concept of extracting training examples from language models, which was later formalized through the notion of $k$-extractability \citep{carlini2021extracting}. A sequence is considered $k$-extractable (or memorized) if, when prompted with $k$ prior tokens from the training data, the model generates the exact continuation that appears in its training corpus. One of the measures of $k$-extractibility is memorization score. The memorization score \citep{chen2024memorization, biderman2023pythia} calculates the proportion of matching tokens between the model's greedy generation and the ground truth continuation:
\begin{equation}
    \label{eq:mem_score}
    M(X, Y) = \frac{1}{n}\sum_{i=1}^{n}\mathbf{1}(x_i = y_i)
\end{equation}

where $n$ is the continuation length, $X$ represents the model-generated tokens, and $Y$ represents the ground truth continuation. A score of 1 indicates perfect memorization, while 0 indicates no overlap. However, this metric has its limitations as it only captures exact token-level matches. To address these limitations, researchers have utilized other more robust metrics like BLEU, Levenshtein distance, embedding similarity, etc. \citep{mccoy2023much, ippolito2022preventing, duan2023mapping,peng2023detecting, reimers2019sentence}.

The selection of an appropriate prefix length $k$ is crucial in memorization studies. Small values of $k$ might lead to ambiguous prompts with many valid continuations, while excessively large values might make the task trivial \cite{chen2024memorization}. Most studies utilize prefix lengths between 30-50 tokens \cite{biderman2023emergent, stoehr2023localizing}, balancing these considerations.

Recent interpretability efforts have made progress in understanding the mechanisms behind memorization. \citet{stoehr2023localizing} found that gradients flow differently for memorized versus non-memorized content, with more gradient activity in lower layers for memorized paragraphs. They identified an attention head (specifically, head 2 in layer 1 of GPT-NEO 125M) that focuses on rare tokens and plays a crucial role in memorization. Similarly, \citet{haviv2023understanding} showed that memorized information exhibits distinct patterns, with early layers promoting tokens in the output distribution and upper layers amplifying confidence.

\subsection{Mechanistic Interpretability}

Mechanistic interpretability seeks to reverse-engineer neural network behavior into human readable explanations. 
Circuit framework \citep{olah2020zoom, elhage2021mathematical} is particularly important for understanding transformer architectures. A circuit is defined as the minimal computational subgraph of a model that faithfully reproduces the model's behavior on a given task \cite{wang2022interpretability, hanna2023does, conmy2023towards}. The computational graph consists of nodes (e.g. attention heads and MLPs) connected by edges that specify information flow. A circuit is considered faithful if corrupting all model edges outside the circuit maintains the model's original task performance. This faithfulness property ensures circuits provide reliable explanations compared to other interpretability methods that may capture features unused by the model \cite{olah2020zoom, elhage2021mathematical}.
Automatic circuit discovery methods have emerged to identify these minimal computational subgraphs efficiently. These include techniques like ACDC \cite{conmy2023towards}, which automate and accelerate the process of finding circuits through systematic interventions. However, as model size grows, the number of required forward passes makes these methods computationally expensive.

Activation patching---also known as causal tracing or interchange intervention---serves as a fundamental technique in circuit discovery, replacing specific internal activations with cached activations from a different input to observe effects on model output \citep{vig2020causal, geiger2021causal, meng2022locating}. This method relies on contrastive datasets with clean and corrupted examples designed to elicit different model behaviors. Researchers have applied activation patching to various tasks ranging from indirect object identification (IOI) \citep{wang2022interpretability}, where models must predict the recipient of an action (e.g., "John gave a bottle to [Mary]"), to factual knowledge retrieval like Capital-Country tasks (e.g., "Vienna is the capital of [Austria]") \citep{meng2022locating}.

Gradient-based methods have greatly improved circuit discovery efficiency. Edge Attribution Patching (EAP) \cite{nanda2023attribution, syed2023attribution} allowed to perform activation patching at scale by using gradients to approximate patching effects, reducing computational requirements from thousands of forward passes to just two forward passes and one backward pass for all possible interventions. 
Recently, Edge Attribution Patching with Integrated Gradients (EAP-IG) \citep{hanna2023edge} addressed the zero gradient problem encountered in EAP by accumulating gradients along the path from corrupted to clean inputs. 
EAP-IG improves the measurement of circuit faithfulness, a concept previously established in circuit analysis \cite{wang2022interpretability}. \citet{miller2024autocircuit} demonstrate that faithfulness metrics are highly sensitive to experimental choices in ablation methodology such as component type, ablation value, token positions, and direction. Their work shows that circuit faithfulness metrics vary significantly across methods, indicating that optimal circuits depend on both task prompts and evaluation methodology. They also released AutoCircuit, an efficient library for circuit discovery \cite{miller2024autocircuit}, which we use in our experiments.

\begin{figure*}[t]
    \centering
    \includegraphics[width=\textwidth]{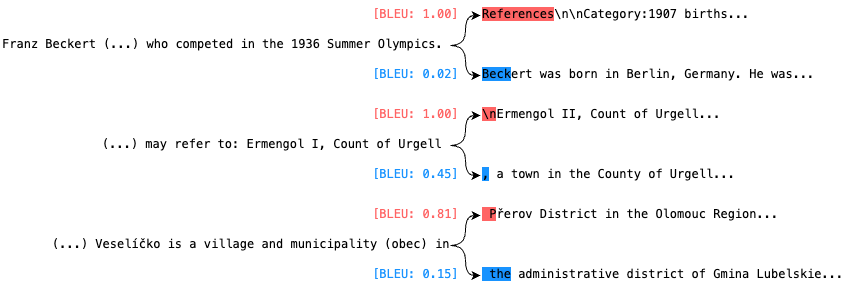}
    \caption{Examples of divergence points in memorized content. Each pair shows a shortened context (left) followed by two possible continuations: the ground truth (red, top) and model's actual generation (blue, bottom) with their respective BLEU scores. Higher BLEU scores (closer to 1.0) indicate stronger memorization.}
    \label{fig:divergence_example}
\end{figure*}

\section{Methodology}
\label{sec:methodology}

Our analysis focused on Wikipedia subset from the Pile dataset \citep{gao2020pile}. Following  \citet{stoehr2023localizing}, we used a 50-50 token split: the first 50 tokens as context and the next 50 tokens as the target continuation. Resulting dataset contained approximately 16 million examples. Again, following \citet{stoehr2023localizing} we selected GPT-Neo-125m as model under analysis to potentially compare our interpretability experiments.

To identify memorized content, we used memorization score (Eq. \ref{eq:mem_score}). After scoring all samples, we retained only those with memorization scores of exactly 1.0, representing perfect memorization. This created a base of 4047 samples for our contrastive datasets.

\subsection{Divergence points}

Using our memorized samples, we identified divergence points - the specific token positions where model generation shifts from memorization to novel generation. 

Our algorithm methodically shortened the context length until it detected a significant drop in BLEU score, using a threshold of 0.3. We chose BLEU over exact token matching because it measures n-gram overlap between sequences. This makes it less sensitive to minor variations, while still effectively detecting when the model branches away from the memorized continuation path. Through this process, we transformed fully memorized contexts into what we term "Potentially Memorized" (PM) contexts (Fig \ref{fig:divergence_example}) - contexts where the model's highest probability token at the trimmed position no longer follows the memorized continuation path.

This approach enabled us to create two distinct contrastive datasets designed for different analytical tasks: memorization decision (\ref{sec:methodology:mem_decision}) and branch comparison (\ref{sec:methodology:branch_comparison}). See Table \ref{tab:dataset_examples} for examples from the datasets.

\begin{table*}[!htb]
\centering
\begin{tabular}{p{0.63\textwidth}|p{0.12\textwidth}|p{0.15\textwidth}}
\hline
\textbf{Sample Context} & \textbf{Prediction} & \textbf{Ground Truth} \\
\hline
\hline
\multicolumn{3}{c}{\textit{Branch Comparison Dataset}} \\
\hline
\hline
\textbf{Clean:} Fazalur Rehman (born...) is a Pakistani field & \textcolor{red}{" hockey"} & \textcolor{red}{" hockey"} \\
\textbf{Corrupted:} Fazalur Rehman (born...) is a Pakistani politician & " who" & - \\
\hline
\textbf{Clean:} Marek (...) is a Polish weightlifter. & \textcolor{red}{" He"} & \textcolor{red}{" He"} \\
\textbf{Corrupted:} Marek (...) is a Polish weightlifter who & " competed" & - \\
\hline
\hline
\multicolumn{3}{c}{\textit{Memorization Decision Dataset}} \\
\hline
\hline
\textbf{Clean:} Kim Woon-sung (born...) is a South Korean fencer & "who" & \textcolor{red}{"."} \\
\textbf{Corrupted:} Seong Nak-gun (born...) is a South Korean sprinter & \textcolor{red}{"."} & "." \\
\hline
\textbf{Clean:} József Farkas is a (...) wrestler. He competed & " in" & \textcolor{red}{" at"} \\
\textbf{Corrupted:} Istvan Zsolt was a (...) football referee. He officiated & \textcolor{red}{" at"} & " at" \\
\hline
\end{tabular}
\caption{Examples from our contrastive datasets showing clean and corrupted samples with their contexts, model predictions, and ground truth tokens. Memorization tokens are shown in \textcolor{red}{red}. For Branch Comparison corrupted samples, the "-" indicates no ground truth as there is no such completion in the Pile.} 
\label{tab:dataset_examples}
\end{table*}

\subsection{Memorization Decision Dataset}
\label{sec:methodology:mem_decision}

This dataset is designed to identify which model components select tokens that lead to memorized continuations.

For the "clean" samples, we preserved PM contexts that are just one token away from triggering either a memorized continuation or a divergent path. For the "corrupted" samples, we selected non-memorized examples from the dataset that would still produce the same token as found in the memorization completion when measured by the model's highest logit value.

The key idea behind this approach is that we need our contrastive pairs to exhibit different behaviors while maintaining semantic similarity. The clean samples sit at the threshold of memorization, while the corrupt samples are maximally distant from memorization when measured by BLEU score. Despite this behavioral difference, we ensured semantic similarity between pairs by calculating embeddings (using the same underlying model) and selecting the closest non-memorized samples. Importantly, only the corrupt samples lead to the memorization token, while clean samples allow model to exhibit its natural behavior. This contrast allows us to isolate the computational mechanisms responsible for exact moment (token position) when model starts memorization.

\subsection{Branch Comparison Dataset}
\label{sec:methodology:branch_comparison}

The question we aim to answer by creating this dataset is: given that model already "decided" to retrieve memorized sentence, how is this behavior different from the model that has branched out? 
For our contrastive pairs, the "clean" examples consist of PM contexts followed by the next token from the memorized sequence. This forces the model to continue along the memorization path. The "corrupted" examples contain the same PM context but are followed by the model's highest probability token prediction, which leads away from memorization.

\subsection{Circuit Discovery}

Formally, a circuit is the minimal computational subgraph of a model whose performance remains close to the whole model's performance on a specific task. For a transformer model with nodes $V$ and edges $E$, a circuit is a subgraph $(V_c, E_c)$ where $V_c \subseteq V$ and $E_c \subseteq E$ that connects input embeddings to output logits. During circuit discovery, we can test the performance impact of different edges by modifying the information flow in the model. For each node $v$ with incoming edges $E_v$, its input is set to:
\begin{equation}
\label{eq:circuit_dag}
\sum_{e=(u,v) \in E_v} i_e \cdot z_u + (1-i_e) \cdot z'u    
\end{equation}
Where $i_e$ indicates if edge $e$ is in the circuit, $z_u$ is node $u$'s output on clean inputs, and $z'_u$ is its output on corrupted inputs. This intervention can be applied to both circuit and non-circuit edges to understand their relative contribution to the task performance.

We used Edge Attribution Patching with Integrated Gradients (EAP-IG) \citep{hanna2023edge} as our circuit discovery method. This method leverages gradients to efficiently approximate the effect of patching specific model components without requiring multiple forward passes. For each edge (u,v) in the model's computational graph, EAP-IG calculates an importance score using:
\begin{equation}
\label{eq:eap_ig}
    (z'u - z_u) \frac{1}{m} \sum{k=1}^{m} \frac{\partial L(z' + \frac{k}{m}(z - z'))}{\partial z_v}
\end{equation}
Where $z$ represents token embeddings for clean inputs, $z'$ for corrupted inputs, $L$ is our loss function, and $m$ is the number of steps used to approximate the integral (we used m=5).

In our patching experiments, we considered both noising and denoising approaches. In the noising approach (corrupt → clean patching), we patch activations from corrupted inputs into clean inputs to identify which components break the model's behavior when corrupted. Conversely, in the denoising approach (clean → corrupt patching), we patch activations from clean inputs into corrupted inputs to identify which components restore the model's behavior. 

Our formulated tasks have various target objectives to optimize. To standardize evaluation across these diverse metrics, we followed \citet{hanna2023edge} to convert each task metric to a normalized faithfulness score. It is defined as $(m - b')/(b - b')$, where $m$ is the circuit's performance, $b$ is the whole model's performance on clean inputs, and $b'$ is performance on corrupted inputs. This normalization enables cross-dataset comparison of circuits.

For each task, we first compute importance scores for model edges using EAP-IG. Using binary search over edges, we then identify smallest number of edges that still results in a circuit that can be considered faithful ($\geq 85\%$ of the complete model's performance). 

\begin{table*}[!htb]
\centering
\begin{tabular}{|p{2.2cm}|p{5.5cm}|p{5.5cm}|}
\hline
\textbf{Task} & \textbf{Noising} & \textbf{Denoising} \\
\hline
\textbf{Memorization \newline Decision} & 
\textbf{Interpretation:} Identifies parts that \textit{promote} memorization when corrupted
\newline
\textbf{Metric:} $logit_{mem}$ ($\uparrow$) or \newline $logit_{mem}-logit_{pred}$ ($\uparrow$) & 
\textbf{Interpretation:} Identifies parts that can \textit{restore} normal behavior by removing memorization
\newline
\textbf{Metric:} $logit_{pred}$ ($\uparrow$) or $logprob_{pred}$($\uparrow$)\\
\hline
\textbf{Branch \newline Comparison} & 
\textbf{Interpretation:} Identifies parts that can \textit{remove} memorization when corrupted
\newline
\textbf{Metric:} $logit_{mem}$ ($\downarrow$) or \newline $accuracy_{mem}$ ($\downarrow$) & 
\textbf{Interpretation:} Identifies parts that can \textit{recover} memorization from a divergent path
\newline
\textbf{Metric:} $logit_{mem}$ ($\uparrow$) or $accuracy_{pred}$ ($\downarrow$) \\
\hline
\end{tabular}
\caption{Comparison of Noising and Denoising approaches across datasets. Arrows indicate whether higher ($\uparrow$) or lower ($\downarrow$) values are better for each metric.}
\label{tab:noising_denoising}
\end{table*}

\section{Experiments}
\label{sec:experiments}

We define \textbf{memorization token ($t_{mem}$)} as the token that appears in memorized continuations, i.e. "ground truth" token from the pre-training dataset. \textbf{Predicted token ($t_{pred}$)} refers to the token that the model predicts with highest probability (i.e. the argmax token of an unpatched model). See Table \ref{tab:noising_denoising} for a summary of tasks.

\subsection{Memorization Decision Task}
For the Memorization Decision task, we investigate which components of the model are responsible for determining whether to retrieve memorized content. We used following objectives for finding edge importances via EAP: $L = logit_{mem} - logit_{pred}$ and $L = logit_{mem}$. 

In our noising experiments, we aim to discover which model components, when activated, can trigger memorization in contexts that wouldn't normally produce it. For these experiments, we measure the memorization token logit or the logit difference between memorization and predicted token, as these directly quantify the model's preference for the memorized content over "natural" continuation.

Our denoising approach seeks to identify components that, when cleaned with PM contexts, can induce memorization on arbitrary not-memorized samples. Here, we use the predicted token logit as our metric, as it shows how effectively the circuit can influence the model toward non-memorized predictions. Since it makes sense to compare logits on specific generation steps on the same samples, we make sure to do exactly that. We also measure the logprobability of the predicted token logit to provide a normalized measure that accounts for the overall confidence distribution across all possible tokens.

\subsection{Branch Comparison Task}
In the Branch Comparison task, we examine how the model's computation differs between continuing along a memorized path versus diverging to a novel generation. Unlike the Memorization Decision task, logit differences between specific tokens are less meaningful here since we're comparing different "branches" of generation.
Instead, we use on $L = -logit_{mem} $.

In addition, we used $\textbf{accuracy}_{t}$ as the percentage of samples where a specific token $t$ receives the highest probability from the model.

With noising, we try to isolate components that, when corrupted, cause the model to abandon memorized paths. We measure this effect using the memorization token logit (which should decrease) or its accuracy, as these metrics directly capture the disruption to memorization behavior.

For denoising, the intuition is that circuits found this way, could pull the model back onto memorized paths even after it has begun generating novel content. Our metrics include the memorization token logit (which should increase) and the accuracy of predicted token (which should decrease), as these best reflect the model's shift back toward memorized content. 

\subsection{Circuit Verification}

Circuit discovery methods can be susceptible to misleading conclusions, making verification beyond task-specific metrics essential. While faithfulness indicates that a circuit maintains performance compared to the full model, it does not guarantee that the circuit truly captures the causal relationships involved in memorization \citep{miller2024autocircuit}.

To address these concerns, we additionally perform the following analysis for all discovered circuits:

\begin{enumerate}
    \item \textbf{Random baseline comparison}: Circuit is compared against randomly selected edges. This helps ensure the circuit's performance is not due to chance or to having a sufficient number of parameters regardless of their function. \textit{During our tests all random circuits showed faithfulness close to zero.}    
    
    \item \textbf{Cross-task generalization}: We evaluate whether circuits discovered in one task maintain their functional properties when applied to the other task. This includes testing Memorization Decision circuits on Branch Comparison to reveal shared memorization mechanisms, and applying Branch Comparison circuits to Memorization Decision to determine if mechanisms that remove or recover memorization in one context can influence token-level decisions in another.
    
    \item \textbf{Cross-corpus generalization}: We test circuit performance across different subsets of the Pile beyond Wikipedia to evaluate how memorization mechanisms generalize across diverse text domains.

    \item \textbf{Alternative patching methods}: We test circuits with different patching types, including zero ablation (setting activations to zero), mean over corrupt dataset, and mean over clean dataset. This verifies that circuit performance is good not just because we patch it with specifically crafted counterfactual examples, but it also can perform well under much noisier patches.
\end{enumerate}

\begin{table*}[t]
\centering
\begin{tabular}{|p{5.5cm}|c|c|c|}
\hline
\textbf{Circuit} & \textbf{Edge Count} & \textbf{Edge \%} & \textbf{Faithfulness} \\
\hline
\multicolumn{4}{|c|}{\textbf{Memorization Decision - Noising}} \\
\hline
($L=logit\_diff$, $logit\_diff$) & 141 & 0.43\% & 0.95 \\
($L=logit_{mem}$, $logit\_diff$) & 332 & 1.02\% & 0.89 \\
($L=logit\_diff$, $logit_{mem}$) & 3,769 & 11.60\% & 0.86 \\
($L=logit_{mem}$, $logit_{mem}$) & 1,923 & 5.92\% & 0.89 \\
\hline
\multicolumn{4}{|c|}{\textbf{Memorization Decision - Denoising}} \\
\hline
($L=logit\_diff$, $logit_{pred}$) & 5,614 & 17.28\% & 0.93 \\
($L=logit_{mem}$, $logit_{pred}$) & 1,923 & 5.92\% & 0.94 \\
\hline
\multicolumn{4}{|c|}{\textbf{Branch Comparison - Noising}} \\
\hline
($L=-logit_{mem}$, $logit_{mem}$) & 78 & 0.24\% & 0.96 \\
($L=-logit_{mem}$, $accuracy$) & 14 & 0.04\% & 0.98 \\
\hline
\multicolumn{4}{|c|}{\textbf{Branch Comparison - Denoising}} \\
\hline
($L=-logit_{mem}$, $logit_{mem}$) & 141 & 0.43\% & 0.95 \\
($L=-logit_{mem}$, $accuracy$) & 14 & 0.04\% & 1.00 \\
\hline
\end{tabular}
\caption{Circuit discovery results across different tasks and metrics. Circuits are identified by their loss function and metric used to calculate faithfulness (see Table \ref{tab:noising_denoising}). Edge percentage represents the proportion of edges in the discovered circuit relative to the full model. Faithfulness scores indicate how closely the circuit's performance matches the full model. Logit diff always means $logit_{mem} - logit_{pred}$.}
\label{tab:circuit_results}
\end{table*}

\section{Results}
\label{sec:results}

As shown in Table \ref{tab:circuit_results}, we have identified some compact ($\leq 1\%$ edges) yet functional circuits ($\geq 85\%$ faithful) for both Memorization Decision and Branch Comparison tasks. 
Most notably, the Branch Comparison circuit was remarkably minimal, requiring only 14 edges (0.04\% of the model) which made us skeptical about this circuit reliability. We validate this and other small circuits in our verification stage. Note that we consider circuits with $\geq 5\%$ of edges noisy and therefore do not perform verification.

\subsection{Cross-Task Generalization}
\label{sec:results:cross-task}

During cross task evaluation we found several asymmetries. 
By noising the model's Memorization Decision circuit (141 edges) with Branch Comparison dataset memorization token accuracy dropped from 97\% to 12 \% showed excellent performance with faithfulness (1.00) (Table \ref{tab:mem_on_branch}). 
This indicates that circuits responsible for token-level memorization decisions are also effective at controlling whether the model continues along memorized paths.
The Branch Comparison circuit (14 edges), when applied to Memory Decision, showed moderate performance in the noising setup --- although logit diff between memorized and not-memorized token drops when corrupted, the value of the memorized logit is not promoted. In general, both Branch Comparison circuits with 14 and 78 edges performed poorly on the Memorization Decision dataset, while the largest circuit of 141 edges achieved only 0.7 faithfulness based on logit diff (Table \ref{tab:branch_on_mem}).
This indicates that once model already started producing memorized completion, its' mechanisms are different from the the mechanisms behind initial "decision".

\subsection{Cross-Subset Generalization}
\label{sec:results:cross-corpus}

For our cross-corpus generalization experiments, we focused on the Branch Comparison task across other subsets of the Pile — specifically GitHub, Enron Emails, and Common Crawl. Due to computational resource constraints, we did not extend these tests to the Memorization Decision task. We tried both noising and denoising approaches, i.e. patching circuits to remove memorization and to induce memorization respectively, see Table \ref{tab:cross-subsets}.

The Memorization Decision circuit effectively reduced memorization token accuracy from 77.6\% to 9.3\% on GitHub and from 90.5\% to 5.2\% on Common Crawl when applied in noising setup.
The Branch Comparison circuit demonstrated similar effectiveness, reducing memorization token accuracy from 77.6\% to 10.2\% on GitHub, 86.4\% to 20.5\% on Enron Emails, and 90.5\% to 4.5\% on Common Crawl.

In denoising experiments, both circuits showed less capabilities in cross-corpus transfer. While faithfulness scores remained high, the actual numeric improvements were less dramatic - for example, the Branch Comparison circuit increased memorization accuracy from 5.2\% to only 23.8\% on Common Crawl. One hypothesis here is that memorization prevention mechanisms may operate by simply promoting alternative tokens rather than specifically targeting memorization, while memorization induction appears more context-dependent, with different datasets likely triggering distinct mechanisms that we collectively identify as memorization behavior.

\subsection{Alternative Patching Methods}

We tested different patching strategies to verify circuit robustness beyond our primary patching approach (Table \ref{tab:ablations}). Memorization Decision circuit showed poor faithfulness with zero and mean-over-dataset ablations, both in noising and denoising settings. This indicates that the circuit's ability to affect normal behavior heavily relies on the crafted counterfactuals.

Similarly, the detailed metrics for the Branch Comparison circuit revealed extremely low token accuracies (0-0.15\%) across all settings, suggesting high dependency to clean-corrupt pairs.

This behavior is, in general, expected as alternative patching methods often produce activations that are too out-of-distribution for the model to process normally \citep{chan2022causal}.

\section{Conclusion}
In this work, we identified and analyzed circuits responsible for memorization behaviors in language models through targeted circuit discovery methods.

The Branch Comparison task appeared easier to disentangle than the Memorization Decision task, requiring only 14 edges (0.04\% of the model) compared to 141 edges (0.43\%) for similar faithfulness levels. However, we should be cautious about this finding, as our results in Section \ref{sec:results} show that these small circuits have limited generalizability across tasks. The extremely small circuit size raises questions about whether we've captured the complete memorization mechanism or merely identified a critical but incomplete component.

Our cross-task generalization results reveal a clear pattern: Memory Decision circuits work well for Branch Comparison tasks, but Branch Comparison circuits perform poorly on Memory Decision tasks. This suggests that Memory Decision circuits contain components that can both detect memorizable content and control its usage, while Branch Comparison circuits mainly handle continuation with less ability to make initial memorization decisions.

Cross-corpus experiments reveal that memorization prevention mechanisms transfer across different datasets, while memorization induction appears more context-dependent. This suggests that different text domains may trigger distinct mechanisms that collectively manifest as memorization behavior, with the ability to prevent memorization being more generalizable than the ability to induce it.

Our experiments show that removing memorization from already-memorized samples (noising approach) is easier than inducing memorization in non-memorized samples (denoising approach). Interestingly, this observation contradicts findings from \citet{stoehr2023localizing}, who demonstrated that memorized continuations are harder to corrupt than non-memorized ones. In their experiments, they showed that even when targeting the top 0.1\% of gradient-implicated weights, memorized passages resisted modification while maintaining their distinctive patterns. 
This difference likely comes from our focus on finding specific computation paths rather than using gradient methods to find important weights. One hypothesis is that memorization works through two different mechanisms: a "trigger" circuit that decides when to use memorized content (which we found), and a more spread-out storage system throughout the model (which they found). The trigger circuit can be easily disrupted, while the stored information itself is harder to change. This suggests that memorization safeguards might work better by targeting these trigger mechanisms rather than trying to remove the stored information.

\section*{Limitations}

Our study faces several methodological limitations worth considering. First, we relied exclusively on Edge Attribution Patching with Integrated Gradients (EAP-IG) to identify circuits. While computationally efficient, attribution patching provides only an approximation of activation patching results. It was shown, that there are cases where attribution patching scores do not fully correlate with direct activation patching measurements \citep{nanda2023attribution,hanna2023edge}, which could affect our circuit identification accuracy. Future work should validate our findings using direct activation patching.

Second, our findings are limited by focusing on a single, relatively small language model (GPT-Neo-125m). Larger models might employ more complex memorization mechanisms or distribute them differently across components. Testing across multiple model sizes and architectures would provide more robust evidence about how memorization mechanisms scale and evolve.

Third, there remains uncertainty about whether our identified circuits represent general memorization mechanisms or are specific to the Wikipedia subset of the Pile or artifacts of our contrastive dataset construction. Despite showing cross-corpus generalization, our datasets were constructed using specific criteria for memorization detection that may not capture all memorization phenomena in language models. Additionally, while we tested generalization across different text domains, all tests were derived from the Pile dataset, potentially limiting the scope of our conclusions.

Future work should investigate how memorization mechanisms scale with model size, whether similar circuits exist in different model architectures, and expand testing to more diverse datasets and memorization criteria to establish the generality of these mechanisms.

\bibliography{custom}

\appendix

\section{Appendix}
\label{sec:appendix}

\begin{table*}[t]
\centering
\begin{tabular}{|l|r|r|r|}
\hline
\textbf{Circuit} & \textbf{accuracy\_mem} & \textbf{accuracy\_pred} & \textbf{faithfulness} \\
\hline
\textit{All edges clean} & 97.35 & \cellcolor{red!15}12.69 & — \\
\hline
\textit{All edges corrupted} & \cellcolor{blue!15}13.26 & 94.5 & — \\
\hline
\rowcolor{gray!15} \multicolumn{4}{|c|}{\textbf{Noising Experiments} - Target: \cellcolor{blue!15}accuracy\_mem = 13.26} \\
\hline
141 edges & \cellcolor{blue!15}12.93 & 61.37 & 1.00 \\
\hline
332 edges & \cellcolor{blue!15}12.93 & 77.71 & 1.00 \\
\hline
\rowcolor{gray!15} \multicolumn{4}{|c|}{\textbf{Denoising Experiments} - Target: \cellcolor{red!15}accuracy\_pred = 12.69} \\
\hline
141 edges & 68.87 & \cellcolor{red!15}13.80 & 0.99 \\
\hline
332 edges & 88.00 & \cellcolor{red!15}13.08 & 1.00 \\
\hline
\end{tabular}
\caption{Results of applying Memorization Decision circuits to the Branch Comparison task. For noising experiments, success is measured by accuracy\_mem approaching the corrupted value (13.26\%); for denoising --- accuracy\_pred approaching the clean value (12.69\%).}
\label{tab:mem_on_branch}
\end{table*}

\subsection{Cross Task Results}
Table \ref{tab:branch_on_mem} shows the performance of Branch Comparison circuits when applied to the Memorization Decision task.
In noising experiments, we want the circuit to shift logit\_diff toward the corrupted value (4.32), with the 141-edge circuit achieving the best performance (2.77, faithfulness 0.73). The 14-edge circuit shows moderate transfer (0.98, faithfulness 0.41), suggesting limited capacity to block memorization decisions.
For denoising, success is measured by how closely logit\_other approaches the clean value (13.02). All circuits perform poorly here, with even the largest 141-edge circuit reaching only 11.35 (faithfulness 0.53), indicating that mechanisms for maintaining memorized paths transfer poorly to inducing memorization decisions.

Table \ref{tab:mem_on_branch} shows the performance of Memorization Decision circuits when applied to the Branch Comparison task.
In noising experiments, we aim for accuracy\_mem to match the clean value (12.69\%). Both the 141-edge and 332-edge circuits achieve near-perfect transfer (12.93\%, faithfulness 1.00), demonstrating that circuits responsible for memorization decisions effectively prevent branch memorization.
For denoising, success is measured by how closely accuracy\_other approaches the corrupted value (13.26\%). Both circuits perform exceptionally well, with the 332-edge circuit achieving near-perfect transfer (13.08\%, faithfulness 1.00), indicating robust generalization of memorization induction mechanisms.

\begin{table*}[t]
\centering
\begin{tabular}{|l|r|r|r|r|r|}
\hline
\textbf{Circuit} & \textbf{logit\_diff} & \textbf{logit\_mem} & \textbf{logit\_pred} & \textbf{logprob\_pred} & \textbf{faithfulness} \\
\hline
\textit{All edges clean} & -1.36 & 11.66 & \cellcolor{blue!15}13.02 & -5.55 & — \\
\hline
\textit{All edges corrupted} & \cellcolor{red!15}4.32 & 13.74 & 9.4 & -9.39 & — \\
\hline
\rowcolor{gray!15} \multicolumn{6}{|c|}{\textbf{Noising Experiments} - Target: \cellcolor{red!15}logit\_diff = 4.32} \\
\hline
14 edges & \cellcolor{red!15}0.98 & 8.68 & 7.70 & -21.82 & 0.41 \\
\hline
78 edges & \cellcolor{red!15}2.02 & 10.30 & 8.28 & -12.60 & 0.59 \\
\hline
141 edges & \cellcolor{red!15}2.77 & 11.78 & 9.01 & -11.42 & 0.73 \\
\hline
\rowcolor{gray!15} \multicolumn{6}{|c|}{\textbf{Denoising Experiments} - Target: \cellcolor{blue!15}logit\_pred = 13.02} \\
\hline
14 edges & -0.11 & 7.95 & \cellcolor{blue!15}8.06 & -20.90 & 0.38 \\
\hline
78 edges & -1.38 & 9.41 & \cellcolor{blue!15}10.79 & -10.99 & 0.38 \\
\hline
141 edges & -1.10 & 10.25 & \cellcolor{blue!15}11.35 & -9.42 & 0.53 \\
\hline
\end{tabular}
\caption{Results of applying Branch Comparison circuits to the Memorization Decision task. For noising experiments, success is measured by logit\_diff approaching the corrupted value (4.32); for denoising --- logit\_pred approaching the clean value (13.02).}
\label{tab:branch_on_mem}
\end{table*}

\subsection{Cross subset results}

Table \ref{tab:cross-subsets} shows the performance of both Memorization Decision and Branch Comparison circuits when applied to other subsets of The Pile dataset (GitHub, Enron Emails, and Common Crawl).

In noising experiments, we measure success by how closely the accuracy\_mem matches the corrupted values (9.87\%, 19.55\%, and 5.24\% for the respective datasets). The Branch Comparison circuit (14 edges) demonstrates remarkable transfer across all subsets, with accuracy values of 10.17\% for GitHub, 20.47\% for Emails, and 4.46\% for Common Crawl, showing very close alignment with target values. The Memorization Decision circuit (141 edges) performs equally well, with particularly strong performance on Emails where it reduces accuracy to 3.28\%, even surpassing the target corrupt value.

For denoising experiments, we want accuracy\_gt to approach the clean values (77.59\%, 86.35\%, and 90.48\%). Both circuits show more modest gains here, with the Branch Comparison circuit increasing accuracy to 12.81\% for GitHub and 8.15\% for Common Crawl, while the Memory Decision circuit achieves similar results with 10.68\% for GitHub, 22.84\% for Emails, and 7.74\% for Common Crawl. These results suggest that while circuits can effectively transfer across datasets to prevent memorization, inducing memorization in non-memorized samples proves more challenging and context-dependent.

\begin{table*}[h]
\centering
\begin{tabular}{|l|c|c|c|c|c|c|}
\hline
\textbf{Dataset/Circuit} & \multicolumn{2}{c|}{\textbf{GitHub}} & \multicolumn{2}{c|}{\textbf{Emails}} & \multicolumn{2}{c|}{\textbf{CC}} \\
\hline
& acc\_mem & acc\_pred & acc\_mem & acc\_pred & acc\_mem & acc\_pred \\
\hline
\textit{All edges clean} & 77.57 & \cellcolor{blue!15}10.89 & 86.41 & \cellcolor{blue!15}20.66 & 90.54 & \cellcolor{blue!15}4.82 \\
\hline
\textit{All edges corrupted} & \cellcolor{red!15}9.87 & 85.49 & \cellcolor{red!15}19.55 & 79.63 & \cellcolor{red!15}5.24 & 82.32 \\
\hline
\rowcolor{gray!15} \multicolumn{7}{|c|}{\textbf{Noising Experiments} - Target: \cellcolor{red!15}acc\_mem close to corrupted values} \\
\hline
Branch Comparison (14 edges) & \cellcolor{red!15}10.17 & 35.96 & \cellcolor{red!15}20.47 & 60.42 & \cellcolor{red!15}4.46 & 17.98 \\
\hline
Mem Decision (141 edges) & \cellcolor{red!15}9.25 & 58.02 & \cellcolor{red!15}3.28 & 57.29 & \cellcolor{red!15}5.24 & 44.64 \\
\hline
\rowcolor{gray!15} \multicolumn{7}{|c|}{\textbf{Denoising Experiments} - Target: \cellcolor{blue!15}acc\_pred close to clean values} \\
\hline
Branch Comparison (14 edges) & 31.49 & \cellcolor{blue!15}12.81 & 22.84 & \cellcolor{blue!15}14.78 & 23.81 & \cellcolor{blue!15}8.15 \\
\hline
Mem Decision (141 edges) & 50.04 & \cellcolor{blue!15}10.68 & 51.53 & \cellcolor{blue!15}22.84 & 47.38 & \cellcolor{blue!15}7.74 \\
\hline
\end{tabular}
\caption{Results of applying memorization circuits to other Pile subsets (Branch Comparison task). The table shows both accuracy of memorized token (acc\_mem) and accuracy of predicted token (acc\_pred) for each dataset and circuit. }
\label{tab:cross-subsets}
\end{table*}

\subsection{Ablation methods results}

Table \ref{tab:ablations} presents the results of our circuit verification using different ablation techniques. We evaluated the Memorization Decision and Branch Decision circuits using three alternative patching methods: Zero ablation (replacing activations with zeros), Mean over clean (replacing with mean values from clean samples), and Mean over corrupt (replacing with mean values from corrupted samples). Each circuit was evaluated on its own respective dataset.

For the Memorization Decision circuit, we measured logit\_diff in noising experiments and logit\_gt in denoising experiments. For the Branch Decision circuit, we measured accuracy\_mem in noising experiments and accuracy\_other in denoising experiments. The "Patching" column represents our default ablation approach using contrastive datasets.

\begin{table*}[h]
\centering
\begin{tabular}{|l|r|r|r|r|}
\hline
\textbf{Circuit \& Metric} & \textbf{Patching} & \textbf{Zero} & \textbf{Mean over clean} & \textbf{Mean over corrupt} \\
\hline
\rowcolor{gray!10} \multicolumn{5}{|l|}{\textit{Memorization Decision Circuit}} \\
\hline
Noising (logit\_diff) & 4.1 & 1.04 & 2.36 & 2.22 \\
\hline
Denoising (logit\_pred) & 7.6 & 10.70 & -1.05 & -1.22 \\
\hline
\rowcolor{gray!10} \multicolumn{5}{|l|}{\textit{Branch Comparison Circuit}} \\
\hline
Noising (accuracy\_mem) & 11.25 & 0.00 & 0.05 & 0.05 \\
\hline
Denoising (accuracy\_pred) & 8.5 & 0.10 & 0.15 & 0.15 \\
\hline
\end{tabular}
\caption{Results of verifying circuits using different ablation techniques}
\label{tab:ablations}
\end{table*}

\end{document}